\pdfoutput=1

\documentclass[11pt]{article}

\usepackage[]{acl}

\usepackage{times}
\usepackage{latexsym}

\usepackage[T1]{fontenc}

\usepackage[utf8]{inputenc}

\usepackage{microtype}

\usepackage{inconsolata}

\usepackage{graphicx}
\usepackage{xcolor}
\usepackage{amsmath}
\usepackage{amssymb}
\usepackage[flushleft]{threeparttable} 
\usepackage{kotex}
\usepackage{booktabs}
\usepackage{makecell}

\newcommand{\Ours}[0]{\textsc{No-codE tool for STatistical analysis of LEgal corpus}}
\newcommand{\ours}[0]{\textsc{Nestle}}

\newcommand{\oururl}[1]{\url{http://nestle-demo.lbox.kr}}
\newcommand{\ourrepo}[1]{{\url{https://github.com/lbox-kr/nestle}}}
\newcommand{\ourvideo}[1]{{\url{https://youtu.be/twkpjYJrvI8}}}

\newcommand{\isla}[0]{\textsc{Isla}}

\newcommand{\embz}{\textsc{Embezzlement}}
\newcommand{\fraud}{\textsc{Fraud}}

\newcommand{\drunk}{\textsc{Drunk driving}}
\newcommand{\rcriminal}{\textsc{Ruling-criminal}}

\newcommand{\korie}[0]{\textsc{KorPrec-IE}}
\newcommand{\lboxopen}[0]{\textsc{LBoxOpen}}
\newcommand{\ljpfacts}[0]{\textsc{LBoxOpen-IE}}

\newcommand{\lexglue}[0]{\textsc{LexGLUE}}

\title{\ours: a No-Code Tool for Statistical Analysis of Legal Corpus}

\author{
Kyoungyeon Cho$^{1,}$\thanks{\quad Equal contribution.} \quad  
Seungkum Han$^{1}$ \quad
Young Rok Choi$^{1}$ \quad
Wonseok Hwang$^{1,2,*,}$\thanks{\quad Corresponding author}\\
  $^1$LBox \quad $^2$University of Seoul\\
  \texttt{\{kycho, hsk2950, yrchoi, wonseok.hwang\}@lbox.kr}
}
\begin{document}
\maketitle
\begin{abstract}
The statistical analysis of large scale legal corpus can provide valuable legal insights.
For such analysis one needs to  
(1) select a subset of the corpus using document retrieval tools, 
(2) structure text using information extraction (IE) systems,
and (3) visualize the data for the statistical analysis.
Each process demands either specialized tools or programming skills whereas no comprehensive unified ``no-code'' tools have been available.
Here we provide \ours, a no-code tool for large-scale statistical analysis of legal corpus.
Powered by a Large Language Model (LLM) and the internal custom end-to-end IE system, \ours\ can extract any type of information that has not been predefined in the IE system opening up the possibility of unlimited customizable statistical analysis of the corpus without writing a single line of code. 
We validate our system on 15 Korean precedent IE tasks and 3 legal text classification tasks from \lexglue.
The comprehensive experiments reveal \ours\ can achieve GPT-4 comparable performance by training the internal IE module with 4 human-labeled, and 192 LLM-labeled examples.

\end{abstract}

\section{Introduction}

\begin{figure}[t!]
\centering
\includegraphics[width=0.45\textwidth]{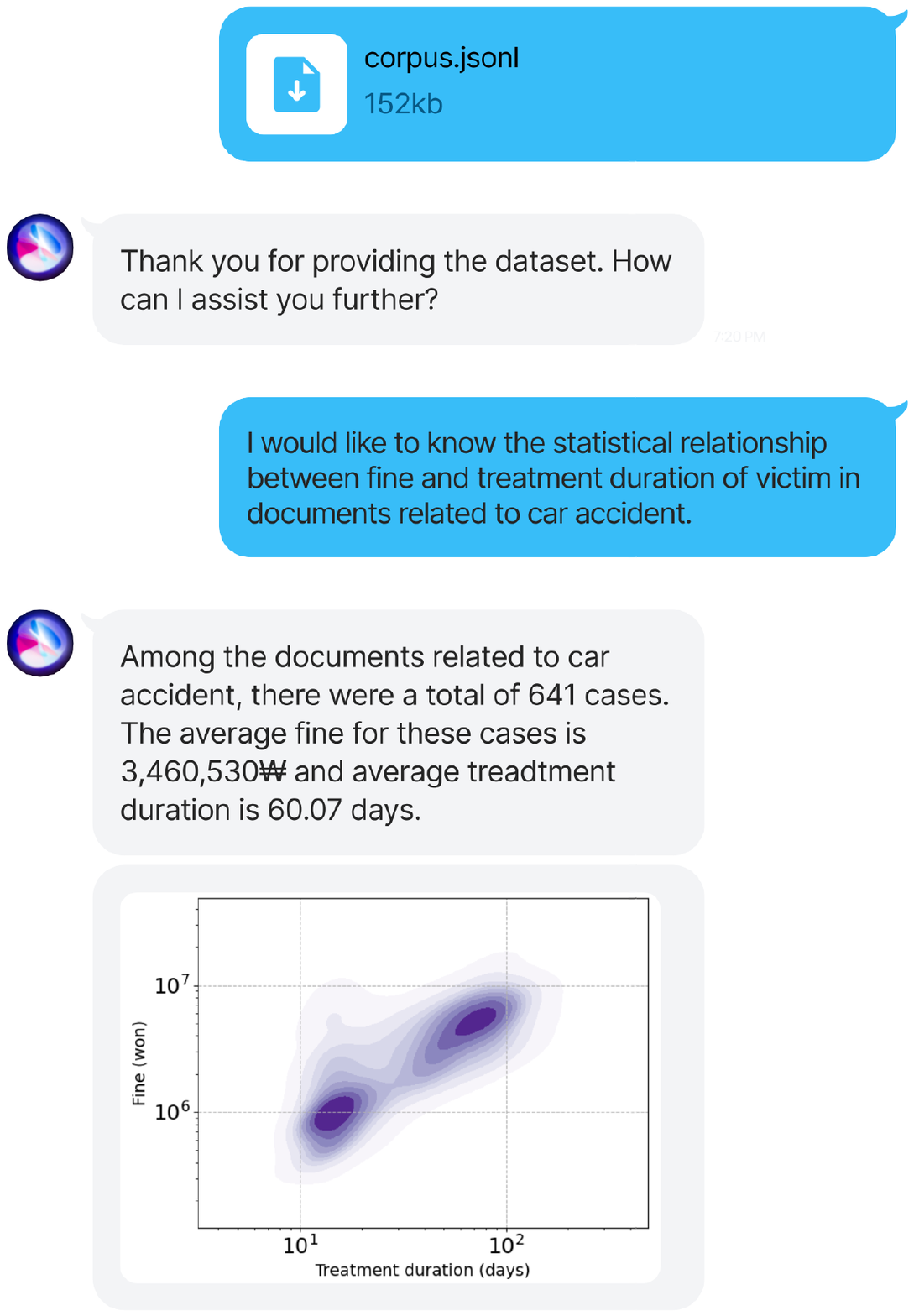}
\caption{Illustration of \ours.}
\label{fig_illu}
\end{figure}

Legal documents include a variety of semi-structured information stemming from diverse social disputes.
For instance, precedents include factual information (such as blood alcohol level in a driving under the influence (DUI) case or loss in an indemnification case) as well as a decision from the court (fine, imprisonment period, money claimed by the plaintiff,  money approved by the court, etc).
While each document contains detailed information about specific legal events among a few individuals, 
community-level insights can be derived only by analyzing a substantial collection of these documents. 
For instance, the consequence of the subtle modification to the statute might only become evident through a comprehensive statistical analysis of the related legal corpus.
Indeed a recent study shows that how the revision of the Road Traffic Act has changed the average imprisonment period in drunk driving cases by analyzing 24k Korean precedents \cite{hwang2022nllp}.

\begin{figure*}[t!]

  \includegraphics[width=1.0\textwidth]{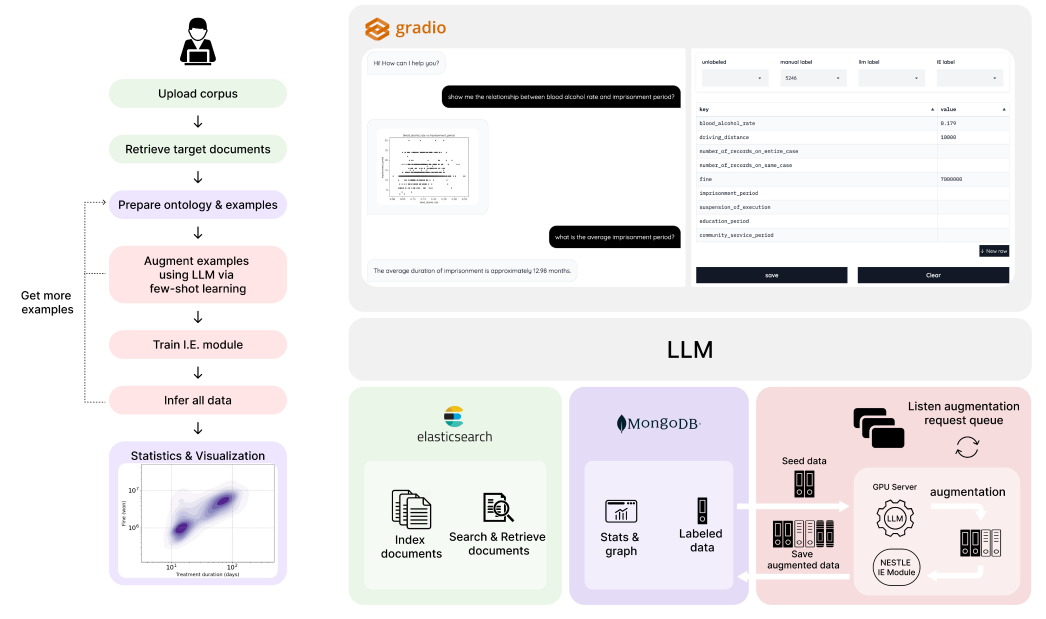}
  \caption{The workflow of \ours}
   \label{fig_arch}
\end{figure*}

Conducting a comprehensive statistical analysis on a legal corpus on a large scale may entail following three key steps: 
(1) choosing a subset of the corpus using retrieval tools, 
(2) structuralizing the documents using information extraction (IE) systems,
and (3) visualizing the data for the statistical analysis.
Each step requires either specialized tools or programming knowledge, impeding analysis for the majority of legal practitioners.
Particularly during text structuration, if the target information is not predefined in the ontology of the IE system, one needs to build their own system.

To overcome such limitation, we developed \ours\footnote{\Ours}, a no-code tool for statistical analysis of legal corpus.
With \ours, users can search target documents, extract information, and visualize statistical information of the structured data via the chat interface, accompanied by an auxiliary GUI for the fine-level controls, such as hyperparameter selection, ontology modification, data labeling, etc.
A unique design choice of \ours\ is the combination of LLM and an custom end-to-end IE system \citep{hwang2022nllp} that brings the following merits. 
First, \ours\ can handle custom ontology provided by users thanks to the end-to-end (generative) property of the IE module.
Second, \ours\ can extract target information from the corpus with as few as 4 examples powered by the LLM. For given few examples, LLM builds the training dataset for the IE module under few-shot setting.
Finally, the overall cost can be reduced by 200 times, and the inference time can be accelerated by 6 times compared to IE systems that rely exclusively on LLM, like ChatGPT, when analyzing 1 million documents.

We validate \ours\ on three legal AI tasks: 
(1) 4 Korean Legal IE tasks \citep{hwang2022nllp}, 
(2) 11 new Korean Legal IE tasks derived from \lboxopen\ dataset \citep{hwang2022lboxopen}, 
and (3) 3 English legal text classification tasks from \lexglue\ \citep{chalkidis2022acl_lexglue, chalkidis-2023-lexglue-chatgpt,tuggener2020ledgar,lippi2018UnfairTOS}.
The comprehensive experiments reveal \ours\ can achieves GPT-4 comparable performance with just 4 human-labeled, and 192 LLM-labeled examples.
In summary, our contributions are as below.
\begin{itemize}
    \item We develop \ours, a no-code tool that can assist users to perform large scale statistical analysis of legal corpus from a few (4--8) given examples.
    \item We extensively validate \ours\ on 15 Korean precedent IE tasks and 3 English legal text classification while focusing on three real-world metrics: accuracy, cost, and time\footnote{The demo is available from \oururl~. The part of the datasets (including 550 manually curated test set for few-shot IE tasks) will be available from \ourrepo.}.
    \item We show \ours\ can achieve GPT-4 comparable accuracy but with 200 times lower cost and in six times faster inference compared to IE systems that solely rely on commercial LLM like ChatGPT for analyzing 1 million  documents.
\end{itemize}

\section{Related Works}

\paragraph{Large Language Model as an Agent}
With rapid popularization of LLM \cite{openai2023gpt4,touvron2023llama,touvron2023llama2,anil2023palm2,claude2023,alpaca2023,zheng2023vicuna}, many recent studies examine the capability of LLM as an agent that can utilize external tools \cite{liang2023taskmatrixai,li2023apibank,liu2023agentbench,wang2023voyager,song2023restgpt,zhuang2023toolqa,tang2023toolalpaca,patil2023gorilla, qin2023toollearning_llm,viswanathan2023prompt2model}.
There are few studies focusing on the capability of LLM as a data analysis agent.
\citeauthor{zhang2023datacopilot} develop Data-Copilot that can help users to interact with various data sources via chat interface.
\citeauthor{ma2023llm_notgood_at_ie} examines the capability of GPT-3 (CODEX, \texttt{code-davinci-002}) as few-shot information extractor on eight NER and relation extraction tasks and propose using LLM to rerank outputs from small language models.
\citeauthor{ding2023acl_gpt3_as_data_anno} evaluate the capability of GPT-3 as a data annotators on SST2 text classification task and CrossNER tasks reporting that GPT-3 shows good performance on SST2.
\citeauthor{he2023annollm} propose `explain-then-annotate' framework to enhance LLM's annotation capability. Under their approach, GPT-3.5 achieves either super-human or human-comparable scores on three binary classification tasks.

Our work is different from these previous works in that we focus on building a no-code tool for ``statistical analysis'' of ``corpus'' where efficient, accurate, yet customizable methods of structuralization of large-scale documents are necessary. Our work is also different in that we focus on information extraction tasks from legal texts.
Finally, rather than performing all IE via LLM, we focus on hybridization between commercial LLM and open-sourced small language model (SLM) by distilling knowledge of LLM to SLM. In this way, the API cost of using LLM does not increase linearly with the size of corpus enabling \ours\ to be applied to industrial scale corpus.

\citeauthor{viswanathan2023prompt2model} recently proposes Prompt2Model allowing users to construct an NLP system by providing a few examples. Compared to Prompt2Model, \ours\ is specialized in large-scale IE task in legal domain and provides additional features like chat-based statistical analysis and GUI for fine-level control. Also \ours\ is rigorously validated on a variety of legal IE tasks.

\paragraph{Information Extraction from Legal Texts}
Previous studies build IE systems for legal texts using tagging-based methods   \cite{cardellino2017eacl_nerc_ontology,mistica2020ie_sentence_classification,hendrycks2021cuad,habernal2022argument_mining,chen2020-joint-entity-relation,pham2021nllp_legal_termolator,hong2021nllu_ie_dialogue,yao2022FACLlevenEventDetection} 
or generative methods \cite{pires2022s2s_ie,hwang2022nllp}.

Our system is similar to \cite{hwang2022nllp} in that we use an end-to-end IE system and focus on statistical analysis of legal information. However our work is unique in that we present a no-code tool and explore hybridization of commercial LLM and open-sourced SLM to expand the scope of analysis to a large-scale corpus while focusing on three real-world metrics: accuracy, time, and  cost.
\begingroup
\setlength{\tabcolsep}{3pt} 
\renewcommand{\arraystretch}{0.5} 
\begin{table*}[t!]
\scriptsize
  \caption{Performance of various models on \korie\ task showing the $F_1$ scores for individual fields: BAC (blood alcohol level), Dist (travel distance), Vehicle (vehicle type), Rec (previous drunk driving record), Loss, Loss-A (aiding and abetting losses), Fine (fine amount), Imp (imprisonment type and period), Susp (execution suspension period), Educ (education period), Comm (community service period). The average scores (AVG) are calculated excluding \drunk\ task, as all models achieve high scores on it.
  Scores are based on test sets, each containing 100 examples per task.
  }
  \label{tbl_comp_korie}
  \centering
  \begin{threeparttable}
  \begin{tabular}{lcccc|c|ccc|c|cc|ccccc}
    \toprule
    \multicolumn{1}{c}{Name} &
    \multicolumn{1}{c}{\makecell{LLM\\module}} &
    \multicolumn{1}{c}{\makecell{IE module\\backbone\\size}} &
    \multicolumn{1}{c}{\makecell{\# of\\training\\examples}} &
    \multicolumn{1}{c}{\makecell{\# of\\LLM-labeled\\examples}} &
    \multicolumn{1}{c}{\makecell{AVG}} &

    \multicolumn{3}{c}{\makecell{\drunk}} & 
    \multicolumn{1}{c}{\makecell{\textsc{Embz}}} & 
    \multicolumn{2}{c}{\makecell{\fraud}} & 
    \multicolumn{5}{c}{\makecell{\rcriminal}}   
    \\
    \midrule
      &   & & (per task) & (per task)
      & $F_1$
      & BAC & Dist & Rec 
      & Loss & Loss & Loss-A 
      & Fine & Imp & Susp & Educ & Comm 
      \\
    \midrule
    \midrule 
    mt5-small$^a$ & - & 0.3B   & 50 & -
    & 58.0
    & 95.8 & 93.0   & 90.1 
    & 72.2 & 42.9   &  0
    & 79.4 & 89.4 & 85.7 & 60.4 & 34.1 
    
    \\
    mt5-large$^a$ & - & 1.2B   & 50 & -
    & 63.9
    & 98.0 & 96.4   & 93.6 
    & 87.5 & 64.8   &  0
    & 84.7 & 82.1 & 96.7 & 68.1 & 27.0
    \\
    \midrule 
    \ours-S$_0$ & ChatGPT & 0.3B    &4$^b$ & 92
    & 62.2
    &  98.0 & 95.3  & 93.0 
    & 70.1 &52.2 & 0.0  
    & 71.2 & 96.5  & 93.6 & 76.7 & 37.5 
    \\
    \ours-S & ChatGPT & 0.3B    &4 & 192
    & 64.7
    &  98.0  & 95.3   & 89.8 
    & 77.3 & 56.5& 0.0 
    & 77.4 & 96.5 & 98.9  & 57.1 & 54.2 
    \\
    \midrule
    \ours-L$_0$ & ChatGPT & 1.2B   &4 & 92
    & 71.8
    & 97.4  & 94.7  & 93.0 
    & 84.9 &65.3 & 0.0 
    & 86.7 & 97.9 & 98.9 & 82.4 & 57.9 
    \\
    \ours-L & ChatGPT & 1.2B   &4 & 192
    & 77.3
    &  98.0 & 95.3  & 91.7
    & 87.0 &68.0 & 11.8  
    & 88.9  & 97.9 & 97.8 & 94.5 & 72.7 
    \\
    \ours-L+ & GPT-4 & 1.2B   & 4 & 192
    & 83.6
    & -  & - & - 
    & 90.5 & 71.2 & 38.1 
    & 89.2 & 95.8 & 98.9 & 96.4 & 88.9 
    \\
    \ours-XXL+ & GPT-4 & 12.9B   &4 & 192
    & 80.4
    & -  &-  &- 
    & 92.5 & 72.6 & 28.6 
    & 92.3 &  96.6 & 96.8 & 88.9 & 75.0
    \\
    \midrule 
    \\
    
    \midrule 
    ChatGPT  & - &  -  & 4 & -
    & 79.6
    & 99.0  & 95.3 & 95.2
    & 87.5 & 75.2 & 34.8
    & 87.1 & 97.8 & 96.5 & 94.7 &  63.4  
    \\
    ChatGPT + aux. inst.  & - & -   & 4 & -
    & -
    & -  &-  & -
    & - & 75.6 & 41.7 
    & 88.5 & 98.6 & 98.8 & 96.4 &  72.7 
    \\
    GPT-4  & - & -   & 4 & -
    & 88.7
    & 98.5  & 97.8 & 92.1
    & 93.5 & 82.3& 59.3 
    & 93.9 & 97.1 & 98.9 & 92.6 & 92.3 
    \\
    \midrule
    \isla$^a$  & - & 1.2B  & --1,000 & -
    & 90.3
    &  99.5 & 97.4  &99.0 
    & 91.7 & 80.3& 69.6
    & 95.5 & 95.7 & 98.9 & 98.2 & 92.3 
    \\

    \bottomrule
  \end{tabular}
  \begin{tablenotes}[]
  \item $a$: From \cite{hwang2022nllp}.
  \item $b$: 8 examples are used in \rcriminal\ task.
  \end{tablenotes}
  \end{threeparttable}
  \vspace{-4mm}

\end{table*}
\endgroup

\section{System}
\ours\ consists of three major modules: a search engine for document retrieval, a custom end-to-end IE systems, and LLM to provide chat interface and label data.
Through conversations with the LLM, users can search, retrieve, and label data from the corpus. 
After labeling a few retrieved documents, users can structure entire corpus using the IE module. 
After that, users can conduct statistical analysis through the chat interface using the LLM. 
Internally, user queries are converted into executable logical forms to call corresponding tools via the ``function calling'' capability of ChatGPT.
The overall workflow is depicted in Fig. \ref{fig_arch}.

\paragraph{Search Engine}
The search engine selects a portion of the corpus for statistical analysis from given user queries.
Utilizing LLM like ChatGPT, we first extract potential keywords or sentences from user queries, then forward them to the search engine for further refinement and selection.
Elasticsearch is used for handling large volumes of data efficiently.

\paragraph{IE Module} 
To structure documents, users first generate a small set of seed examples via either a chat interface or GUI for fine-level control.
Then LLM employs these seed examples to label other documents via few-shot learning.
The following prompt is used for the labeling
\\
\\
{\scriptsize 
\texttt{
You are a helpful assistant for IE tasks. After reading the following text, extract information about {\it FIELD-1, FIELD-2, ..., FIELD-n} in the following JSON format. '{\it FIELD-1}: [value1, value2, ...], {\it FIELD-2}: [value1, value2, ...], ..., {\it FIELD-n}: [value1, value2, ...]'. 
\\{\it TASK DESCRIPTION}
\\{\it INPUT TEXT 1}, {\it PARSE 1}
\\{\it INPUT TEXT 2}, {\it PARSE 2}
\\...
\\{\it INPUT TEXT n}, {\it PARSE n}
\\{\it INPUT TEXT }\footnote{The original prompt is written in Korean but shown in English for the clarity.
}
}
}
\\

The generated examples are used to train the IE model. We use open-sourced language model multilingual T5 (mt5) \cite{xue2021aclmt5} as a backbone.
mt5 is selected as (1) it provides checkpoints of various scale up to 13B, and (2) previous studies show Transformers with encoder-decoder architecture perform better than decoder-only models in IE tasks \cite{hwang2022nllp,hwang2022lboxopen}. The model has also demonstrated effectiveness in distilling knowledge from LLM for QA tasks \citep{li2022explanations}.
The trained model is used to parse remaining documents retrieved from previous step.

\section{Demo}
In this section, we provide the explanation for our demo. The video is also available at \ourvideo.

\paragraph{Labeling Interface}
Users can upload their data (unstructured corpus) using an upload button. Althernatively, they can test the system with examples prepared from 7 legal domains by selecting them through the chat interface. Each dataset comes with approximately 1500 documents and 20 manually labeled examples. After loading the dataset, users can view and perform manual labeling on documents using the dropdown menu where the values of individual fields (such as blood alcohol level, fine amount, etc.) can be labeled or the new fields can be introduced.
The changes are automatically saved to the database.

\paragraph{IE Module Interface}
Users can select options such as model size, number of training epochs, and number of training examples within the IE Module Interface. 
The training of IE module typically takes from 40 minutes to an hour, depending on the parameters above. 
The data is automatically augmented by LLM when the number of manually labeled examples is less than the specified number of training examples above.

\paragraph{Statistical Analysis Interface}
Using the chat interface from the second tab of our demo, users can perform various statistical analyses such as data visualization and calculation of  various statistics. Users can also retrieve a target document upon request.

\section{Experiments}
All experiments are performed on NVIDIA A6000 GPU except the experiments with mt5-xxl where eight A100 GPUs are used.
The IE module of \ours\ is fine-tuned with batch size 12 with learning rate 4e-4 using AdamW optimizer. 
Under this condition, the training sometimes becomes unstable. In this case, we decrease the learning rate to 3e-4. 
The high learning rate is purposely chosen for the fast training.
The training are stopped after 60 epochs (\ours-S), or after 80 epochs (\ours-L, \ours-L+).
In case of \ours-XXL, the learning rate is set to 2e-4 and the model is trained for 20 epochs with batch size 8 using deepspeed stage 3 offload \cite{ren2021deepspeed_zerooffload}. 
For efficient training, LoRA is employed in all experiments \cite{hu2022lora} using PEFT library from Hugging face \cite{peft2022}.
In all evaluations, the checkpoint from the last epoch is used.

For the data labeling, we use ChatGPT version \texttt{gpt-3.5-turbo-16k-0613} and GPT-4 version \texttt{gpt-4-0613}. In all other operations with LLM, we use the same version of ChatGPT except during normalization of numeric strings such as imprisonment period and fines where \texttt{gpt-3.5-turbo-0613} is used.
We set temperature 0 to minimize the randomness as IE tasks do not require versatile outputs. 
The default values are used for the other hyperparameters.
During the few shot learning, we feed LLM with the examples half of which include all fields defined in the ontology while the remaining half are selected randomly.

\begingroup
\setlength{\tabcolsep}{1pt} 
\renewcommand{\arraystretch}{0.6} 
\begin{table*}[t!]
\scriptsize
  \caption{Performance of GPT-4 and \ours-L on the seven criminal IE tasks from \ljpfacts. $F_1$ scores are shown: nRec (the number of identical criminal records), nRec-A (the number of criminal records), Waiver (the victim's intent to waive punishment), Injury (the extent of injuries), and Gender (the victim's gender).
  }
  \label{tbl_comp_ljp-criminal}
  \centering
  \begin{threeparttable}
  \begin{tabular}{l|c|ccc|cc|cccc|cccc|c|cc|ccc}
    \toprule
    \multicolumn{1}{c}{Name} &
    \multicolumn{1}{c}{AVG} &
    \multicolumn{3}{c}{\makecell{Indecent Act.$^1$}} & 
    \multicolumn{2}{c}{\makecell{Obstruction$^2$}}  &
    \multicolumn{4}{c}{\makecell{Traffic injuries $^3$}}  &
    \multicolumn{4}{c}{\makecell{Drunk driving $^4$}}  &
    \multicolumn{1}{c}{\makecell{Fraud $^5$}}  &
    \multicolumn{2}{c}{\makecell{Injuries $^6$}} &  
    \multicolumn{3}{c}{\makecell{Violence $^7$}}  
    \\
    & $F_1$
    & nRec & nRec-A & Waiver
    & nRec & nRec-A 
    & nRec & nRec-A & Waiver & Injury
    & nRec & nRec-A & BAC & Dist
    & Loss
    & Injury & Gender 
    & nRec& nRec-A & Gender
    \\
    \midrule
    \midrule 
    GPT-4 
    & 81.1
    & 88.2 & 85.7 & 83.1
    & 78.7 & 82.6
    & 55.6 & 66.7 & 68.4 & 96.0
    & 88.2 & 88.2 & 100 & 99.0
    & 94.9
    & 94.1&81.6
    & 47.1& 61.3&81.6
    \\
    \midrule 
    \ours-L
    & 78.1
    & 88.9 & 76.5 & 52.9
    & 71.8& 57.1
    & 73.4& 78.0 & 71.9 & 95.8
    & 71.8 & 64.9 & 100 & 96.9
    & 81.0
    & 96.9 & 75.0
    & 64.9& 71.8& 93.6
    \\
    \bottomrule
  \end{tabular}
  \begin{tablenotes}[]
  \item $1$: Indecent act by compulsion (강제추행), $2$: Obstruction of performance of official duties (공무집행방해), $3$: Bodily injuries from traffic accident (교통사고처리특례법위반(치상), $4$: Drunk driving (도로교통법위반(음주운전)), $5$: Fraud (사기), $6$: Inflicting bodily injuries (상해), $7$: Violence (폭행)
  \end{tablenotes}
  \end{threeparttable}
\end{table*}
\endgroup

\begingroup
\setlength{\tabcolsep}{1pt} 
\renewcommand{\arraystretch}{0.6} 
\begin{table}[]
\scriptsize
  \caption{Performance of GPT-4 and \ours-L on the four civil IE tasks from \ljpfacts. $F_1$ scores for individual fields are shown: Dom (the event domain such as real estate, fire incident, etc), Ctr (the type of contract), Exp (the amount of money that plaintiffs spent), Loan (the sum of money borrowed by the defendant), and Relat (the relation between plaintiff and defendant).
  }
  \label{tbl_comp_ljp-civil}
  \centering
  \begin{threeparttable}
  \begin{tabular}{l|c|ccc|cc|ccc|ccc}
    \toprule
    \multicolumn{1}{c}{Name} &
    \multicolumn{1}{c}{AVG} &
    \multicolumn{3}{c}{\makecell{Indem$^1$}} & 
    \multicolumn{2}{c}{\makecell{Loan$^2$}}  &
    \multicolumn{3}{c}{\makecell{UFP$^3$}}  &
    \multicolumn{3}{c}{\makecell{LFD$^4$}} 
    \\
    & $F_1$
    & Dom & Ctr & Exp
    & Loan & Relat
    & Dom & Ctr  & Relat
    & Dom & Ctr & Relat

    \\
    \midrule
    \midrule 
    GPT-4 
    & 83.1
    & 97.0 & 90.4 & 95.8
    & 73.2 & 93.3
    & 93.9& 64.9&59.4
    & 92.8& 73.9& 79.1

    \\
    \midrule 
    \ours-L 
    & 71.5
    & 73.4 & 63.9 & 82.9
    & 59.2 & 30.5
    & 82.4&78.0 & 83.7
    & 87.4& 64.4 & 81.0

    \\
    \bottomrule
  \end{tabular}
  \begin{tablenotes}[]
  \item $1$: Price of indemnification (구상금), $2$: Loan (대여금), $3$: Unfair profits (부당이득금), $4$: Lawsuit for damages (손해배상(기))
  \end{tablenotes}
  \end{threeparttable}
  \vspace{-4mm}

\end{table}
\endgroup

\section{Results}

We validate \ours\ on 15 Korean precedent IE tasks and 3 English legal text classification tasks. 

15 Korean precedent IE tasks are further divided into two categories: \korie\ which consists of 4 tasks from criminal cases previously studied in \cite{hwang2022nllp}
and \ljpfacts, which is generated from \lboxopen~\citep{hwang2022lboxopen} using the factual descriptions from 7 criminal cases and 4 civil cases.
In all tasks, a model needs to extract a legally important information from factual description or ruling of cases such as blood alcohol level, fraud loss, fine, and imprisonment period, the duration of required hospital treatment for injuries, etc.

Three classification tasks are EURLEX, LEDGAR, and UNFAIR-ToS from \lexglue\ \cite{chalkidis2022acl_lexglue,tuggener2020ledgar,lippi2018UnfairTOS}. EURLEX dataset consists of a pair of European Union legislation (input) and corresponding legal concepts (output) from the EuroVoc Thesaurus. In LEDGAR task, a model needs to classify the paragraphs from contracts originated from US Securities and Exchange Commission fillings. Similarly, UNFAIR-ToS is a task of predicting 8 types of unfair contractual terms for given individual sentences from 50 Terms of Service.
These 3 classification tasks are used to demonstrate \ours\ on common (English) legal AI benchmark and also to show \ours\ can be applied to general AI tasks that can be represented in text-to-text format \cite{raffle2019t5}.

\paragraph{\ours\ shows competent performance with only four examples }
We first validate \ours\ on \korie\ that consists of four tasks: \drunk, \embz, \fraud, and \rcriminal.
With four seed examples and 92 LLM-labeled examples, we train mt5-small \cite{xue2021aclmt5}. The result shows that our method already achieves + 4.2 $F_1$ on average compared to the case trained with 50 manually labeled examples (Table \ref{tbl_comp_korie}, 1st vs 3rd rows, 5th column).

\begin{figure}[t]
\centering
\includegraphics[width=0.45\textwidth]{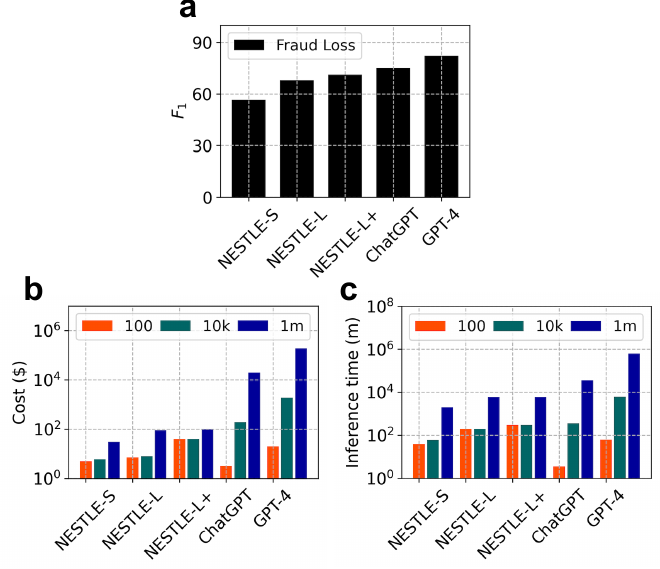}
\caption{Trade-off analysis on \fraud\ task focuses on three real-world metrics: (a) accuracy, (b) cost, and (c) time. } 
\label{fig_trade_off}

\end{figure}

\paragraph{\ours\ can achieve GPT-4 comparable performance}
To enhance the accuracy of \ours, we scale both the quantity of labeled examples by LLM and the size of the backbone of \ours's end-to-end IE module.
With a greater quantity of LLM-labeled examples (from 92 to 192), \ours\ achieves +2.5 $F_1$ on average (3rd vs 4th rows) while the labeling time increases (for example, from 2.4 minutes to 10.6 minutes in \fraud\ task).
With a larger backbone (from mt5-small (0.3B) to mt5-large (1.2B)), \ours's shows  +9.6 $F_1$ (3rd vs 5th rows). 
With both, \ours\ shows +15.1 $F_1$ (3rd vs 6th rows). However, both the labeling time and the training time increase (for example, from 15 minutes to 170 minutes in \fraud\ task).

If the accuracy of teacher model (ChatGPT) is low, the performance of student (mt5) may be bounded by it. 
To check the upper bound of the achievable accuracies, we measure the few-shot performance of ChatGPT. 
\ours-L and ChatGPT shows only 2.3 $F_1$ difference on average (6th vs 9th rows, 5th column) indicating the student models may approach the upper bound. 
To improve \ours\ further, we replace ChatGPT with GPT-4. Although the labeling time and cost increase roughly by 10 times, the average scores increase by +6.3 $F_1$ (Table \ref{tbl_comp_korie} 6th vs 7th rows). 
Notably, this score is higher than ChatGPT by +4.0 $F_1$ (7th vs 9th rows). 

Next we attempt to scale the backbone of the IE module from mt5-large to mt5-xxl (12.9B). Note that unlike commercial LLMs, the IE module can be trained on multiple GPUs for efficient training and indeed the total training time decreases by 70 minutes even compared to a smaller model (\ours-L) by changing GPU from a single A6000 GPU to eight A100 GPUs. However, we could not observe noticeable improvement in $F_1$.

\paragraph{ \ours\ can be generalized to other datasets}
Although we have validated \ours\ on \korie, the dataset mainly consists of numeric fields from criminal cases.
For further validation, we build \ljpfacts\ from \lboxopen~\cite{hwang2022lboxopen}.
\ljpfacts\ consists of 7 tasks from criminal cases (Table \ref{tbl_comp_ljp-criminal}) and 4 tasks from civil cases (Table \ref{tbl_comp_ljp-civil}). 
Compared to \korie, the target fields are more diverse including non-numeric fields such as a contract type, plaintiff and defendant relation, victims' opinion, incident domain, etc as well as numeric fields such as the extent of injury, number of previous criminal records, loan, and more. 

We use \ours-L and measure the performance on manually curated 550 examples (50 for each task). 
\ours-L achieves a GPT-4 comparable performance in 7 criminal tasks (Table \ref{tbl_comp_ljp-criminal}, 78.1 vs 81.1) 
and lower performance in 4 civil tasks (Table \ref{tbl_comp_ljp-civil}, 71.5 vs 83.1). 
This implies \ours\ can be used to glimpse the statistical trend of specific information included in a corpus, but some care must be taken as their accuracies range between $\sim$70 and $\sim$90. 
To overcome this limitation, \ours\ also offers a GUI for rectifying the LLM-augmented examples and collecting more examples manually.
In general, higher accuracy can be achieved by utilizing a specialized backbone in the IE module for the target tasks, alongside a more robust LLM, which is a direction for our future work. 

Finally, the further validation on three English legal text classification tasks from \lexglue\ shows
\ours-L can achieve ChatGPT comparable performance (Table \ref{tbl_comp_lexglue}, 2nd vs 3rd rows).

\begingroup
\setlength{\tabcolsep}{3pt} 
\renewcommand{\arraystretch}{0.6} 
\begin{table}[]
\scriptsize
  \caption{$F_1$ scores of ChatGPT and \ours-L on EURLEX, LEDGAR, and UNFAIR-ToS from \lexglue\ were evaluated using 1,000 random samples from their original test sets, following \cite{chalkidis-2023-lexglue-chatgpt}.
  The number of manually labeled examples ($n_{\text{train}}$) and the number of LLM-labeled examples ($n_{\text{LLM}}$) are shown in the 2nd and 3rd columns respectively..
  }
  \label{tbl_comp_lexglue}
  \centering
  \begin{threeparttable}
  \begin{tabular}{lcc|cc|cc|cc}
    \toprule
    \multicolumn{1}{c}{Name} &
    \multicolumn{1}{c}{\makecell{$n_{\text{train}}$}} &
    \multicolumn{1}{c}{\makecell{$n_{\text{LLM}}$}} &
    \multicolumn{2}{c}{\makecell{EURLEX}} & 
    \multicolumn{2}{c}{\makecell{LEDGAR}} & 
    \multicolumn{2}{c}{\makecell{UNFAIR-ToS}}  
    \\
      &  & 
      & $\mu$-F$_1$ & m-F$_1$
      & $\mu$-F$_1$ & m-F$_1$ 
      & $\mu$-F$_1$ & m-F$_1$ 
      \\
    \midrule
    \midrule
    ChatGPT$^a$ 
    & 8 & -   
    & 24.8 & 13.2
    & 62.1 & 51.1 
    & 64.7 &32.5
    \\
    ChatGPT$^b$ 
    & 32 &  -
    & 33.0 & 18.3
    & 68.3 & 55.6 
    & 88.3  & 57.2
    \\
    \ours-L 
    & 32 & 192
    & 34.1 & 16.7 
    & 58.8& 41.5
    & 91.5 & 51.4
    \\
    \bottomrule
  \end{tabular}
  \begin{tablenotes}[]
  \item $a$: \texttt{gpt-3.5-turbo-0301}. From \cite{chalkidis-2023-lexglue-chatgpt}.
  \item $b$: \texttt{gpt-3.5-turbo-16k-0613}.
  \end{tablenotes}
  \end{threeparttable}
\end{table}
\endgroup

\section{Analysis}
We have shown that \ours\ can extract information with accuracies comparable to GPT-4 on many tasks.
In this section, we extend our comparison of \ours\ to commercial LLMs focusing on two additional real-world metrics: cost and time.
As a case study, we select \fraud\ task from \korie\ where all models struggled (Table \ref{tbl_comp_korie}, 11th and 12th columns, Fig. \ref{fig_trade_off}a). 
We calculate the overall cost by summing up (1) manual labeling cost, (2) API cost, and (3) training and inference cost. 
The manual labeling cost is estimated from the cost of maintaining our own labeling platform (the cost of employing part-time annotators is considered).
The API cost is calculated by counting input and output tokens and using the pricing table from OpenAI. 
The training and inference cost is calculated by converting the training and inference time to dollars based on Lambdalabs GPU cloud pricing.
Note that the API cost increases linearly with the size of the corpus when using commercial LLM.
On the other hand, in \ours, only the inference cost increases linearly with the size of the corpus.
The results show that, for 10k documents, the overall cost of \ours-L is only 4\% of ChatGPT and 0.4\% of GPT-4 (Fig. \ref{fig_trade_off}b).
For 1 million documents, the overall cost of \ours-L is 0.5\% of ChatGPT and 0.05\% of GPT-4 (Fig. \ref{fig_trade_off}b). 
This highlights the efficiency of \ours. 
Similarly, the estimation of overall inference time for 1 million documents reveals \ours-L takes 83\% or 99\% less time compared to ChatGPT or GPT-4 respectively\footnote{The further detailed comparison is available from \ourrepo.}.

\section{Conclusion}
We develop \ours, a no-code tool for statistical analysis of legal corpus. 
To find target corpus, structure them, and visualize the structured data, we combine a search engine, a custom end-to-end IE module, and LLM.
Powered by LLM and the end-to-end IE module, \ours\ enables unrestricted personalized statistical analysis of the corpus. 
We extensively validate \ours\ on 15 Korean precedent IE tasks and 3 English legal text classification tasks while focusing on three real-world metrics: accuracy, time, and cost.
Finally, we want to emphasize that although \ours\ is specialized for legal IE tasks, the tool can be easily generalized to various NLP tasks that can be represented in a text-to-text format.

\section{Ethical Considerations}
The application of legal AI in the real world must be approached  cautiously. Even the arguably most powerful LLM, GPT-4, still exhibits hallucinations \cite{openai2023gpt4} and its performance in the real world legal tasks is still limited \cite{shui2023comprehensive_ljp,zhong2023agieval,martinez2023reeval_bar_exam_gpt4}. 
This may imply that AI systems offering legal conclusions should undergo thorough evaluation prior to being made accessible to individuals lacking legal expertise.

\ours\ is not designed to offer legal advice to general users; instead, it aims to assist legal practitioners by providing statistical data extracted from legal documents.
Furthermore, to demonstrate the extent to which \ours\ can be reliably used for analysis, we conducted extensive validation on 15 IE tasks. 
While \ours\ shows generally high accuracy, our experiments reveal that \ours\ is not infallible, indicating that the resulting statistics should be interpreted with caution.

All the documents used in this study consist of Korean precedents that are redacted by the Korean government following the official protocol \cite{hwang2022lboxopen}.

\section*{Acknowledgements}
We thank Gene Lee for his critical reading of the manuscript, Minjoon Seo for his insightful comments, Paul Koo for his assistant in preparing the figures, and Min Choi for her assistant in preparing the demo.

\bibliography{legal_ai}
\end{document}